\documentclass[letterpaper, 10 pt, conference]{ieeeconf}  % Comment this line out if you need a4paper

\usepackage{xcolor}
\usepackage{pgfplots}
\pgfplotsset{compat=newest}
\usetikzlibrary{svg.path}
\usepackage{siunitx}
\usepackage{amsmath}
\usepackage{hyperref}
\usepackage{multirow}
\usepackage{booktabs}
\usepackage{fixltx2e}
\usepackage{cite}
\usepackage{amssymb}
\usepackage{blindtext}
\usepackage{verbatim}
\usepackage{array}
\usepackage[OT1]{fontenc}
\usepackage{todonotes}
\usepackage{dsfont}
\newcolumntype{L}[1]{>{\raggedright\let\newline\\\arraybackslash\hspace{0pt}}m{#1}}
\newcolumntype{C}[1]{>{\centering\let\newline\\\arraybackslash\hspace{0pt}}m{#1}}
\newcolumntype{R}[1]{>{\raggedleft\let\newline\\\arraybackslash\hspace{0pt}}m{#1}}

\usepackage[linesnumbered,lined,ruled,commentsnumbered]{algorithm2e}
\usetikzlibrary{arrows.meta}

\definecolor{amber}{rgb}{1.0, 0.75, 0.0}
\definecolor{magenta}{rgb}{1.0, 0.0, 1.0}
\definecolor{cyan}{rgb}{0.0, 1.0, 1.0}
\definecolor{c_yellow}{RGB}{248,231,28 }
\definecolor{c_green}{RGB}{126, 211, 33 }
\definecolor{c_purple}{RGB}{189, 16, 224 }
\DeclareSIUnit\px{px}
\DeclareSIUnit\fps{FPS}

\IEEEoverridecommandlockouts                              % This command is only needed if
\title{\LARGE \bf
Single-Shot 3D Detection of Vehicles from Monocular RGB Images via Geometry Constrained Keypoints in Real-Time
}

\author{Nils G{\"a}hlert$^{1}$, Jun-Jun Wan$^{2}$, Nicolas Jourdan$^{3}$, Jan Finkbeiner$^{4}$, Uwe Franke$^{5}$ and Joachim Denzler$^{6}$% <-this % stops a space
\thanks{$^{1}$Mercedes-Benz AG, University of Jena, \newline
        {\tt\small nils.gaehlert@daimler.com}}%
\thanks{$^{2}${Robert Bosch GmbH, \tt\small kuanih.junjun.wan@gmail.com}}%
\thanks{$^{3}$Mercedes-Benz AG \& TU Darmstadt, \newline
        {\tt\small n.jourdan@ptw.tu-darmstadt.de}}%
\thanks{$^{4}$Mercedes-Benz AG,
        {\tt\small jan.finkbeiner@daimler.com}}%
\thanks{$^{5}$Mercedes-Benz AG,
        {\tt\small uwe.franke@daimler.com}}%
\thanks{$^{6}$Computer Vision Group, University of Jena,\newline
        {\tt\small joachim.denzler@uni-jena.de}}%
}

\makeatletter
\let\NAT@parse\undefined
\makeatother

\SetArgSty{textnormal}

\usetikzlibrary{arrows}

\newcommand\copyrighttext{%
  \footnotesize \textcopyright 2020 IEEE. Personal use of this material is permitted. 
  Permission from IEEE must be obtained for all other uses, 
  in any current or future media, including reprinting/republishing 
  this material for advertising or promotional purposes, creating new
  collective works, for resale or redistribution to servers or lists, or 
  reuse of any copyrighted component of this work in other works.
}
\newcommand\copyrightnotice{%
\begin{tikzpicture}[remember picture,overlay]
\node[anchor=south,yshift=10pt] at (current page.south) {\fbox{\parbox{\dimexpr\textwidth-\fboxsep-\fboxrule\relax}{\copyrighttext}}};
\end{tikzpicture}%
}

\begin{document}

\newcommand{\lukas}[1]{{\color{red}{lukas: {#1}}}}
\newcommand{\nils}[1]{{\color{cyan}{nils: {#1}}}}

\newcommand{\netnamepre}{{\color{red}{3D-GCK}}}
\newcommand{\netname}{3D-GCK}

\maketitle

\copyrightnotice

\thispagestyle{empty}
\pagestyle{empty}

%%%%%%%%%%%%%%%%%%%%%%%%%%%%%%%%%%%%%%%%%%%%%%%%%%%%%%%%%%%%%%%%%%%%%%%%%%%%%%%%
\begin{abstract}
In this paper we propose a novel 3D single-shot object detection method for detecting vehicles in monocular RGB images. Our approach lifts 2D detections to 3D space by predicting additional regression and classification parameters and hence keeping the runtime close to pure 2D object detection. The additional parameters are transformed to 3D bounding box keypoints within the network under geometric constraints. Our proposed method features a full 3D description including all three angles of rotation without supervision by any labeled ground truth data for the object's orientation, as it focuses on certain keypoints within the image plane. While our approach can be combined with any modern object detection framework with only little computational overhead, we exemplify the extension of  SSD for the prediction of 3D bounding boxes. We test our approach on different datasets for autonomous driving and evaluate it using the challenging KITTI 3D Object Detection as well as the novel nuScenes Object Detection benchmarks. While we achieve competitive results on both benchmarks we outperform current state-of-the-art methods in terms of speed with more than \SI{20}{\fps} for all tested datasets and image resolutions.
\end{abstract}

%%%%%%%%%%%%%%%%%%%%%%%%%%%%%%%%%%%%%%%%%%%%%%%%%%%%%%%%%%%%%%%%%%%%%%%%%%%%%%%%
\section{INTRODUCTION}
% !TeX root = root.tex

Object detection -- both in 2D as well as in 3D -- is a key enabler for autonomous driving systems. To this end, autonomous vehicles that are currently in development as well as  consumer cars that provide advanced driver assistance systems are equipped with a decent set of sensors such as RGB cameras, LiDARs as well as Radar systems. 
While the accurate distance measurement of LiDAR sensors enables robust 3D bounding box detection, the high cost may prohibit their use in series production vehicles.
3D object detection from monocular RGB cameras thus became a focus in recent computer vision research.
In contrast to LiDAR measurements, RGB images provide rich semantic information that can be used to boost object classification. One of the most challenging problems in 3D object detection from monocular RGB images is the missing depth information. A neural network thus needs to accurately estimate depth from monocular RGB images.
Furthermore, real-time performance is required to enable the use of an algorithm in an autonomous vehicle.

\begin{figure}[t]
\begin{center}
\includegraphics[trim=50 650 50 50, clip, width=0.49\textwidth]{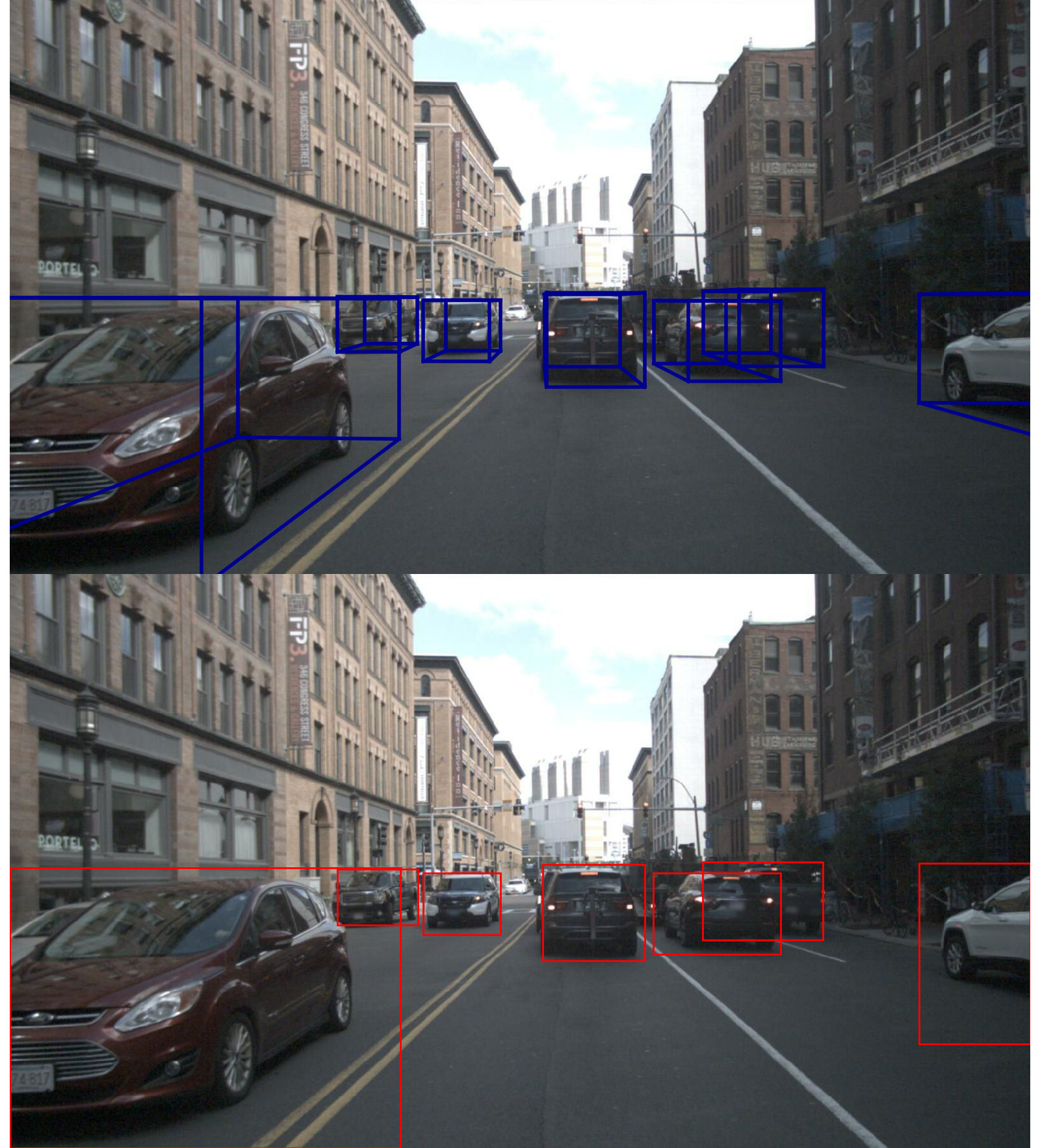} 
\end{center}
\caption{Exemplary result of \netname{} for an image taken from the nuScenes \emph{test} set \cite{caesar2019nuscenes}.  \label{fig:fig1}}
\end{figure}

In this paper we present a novel technique to detect vehicles as 3D bounding boxes from monocular RGB images by transforming a set of predicted regression and classification parameters to geometrically constrain 3D keypoints called \netname. In contrast to other 3D bounding box estimators \netname{} is capable of predicting all three angles of rotation $(\theta, \psi, \phi)$ which are required for a full description of a 3D bounding box. \netname{} focuses only on keypoints in the image plane and exploits the projection properties to generate 3D orientation information. Hence, no labeled ground truth for the angles of rotation is required to train \netname{} which facilitates the collection of training data.

We use a standard single-shot 2D object detection framework -- in our case SSD \cite{liu2016ssd} -- and add the proposed extension to lift the predicted 2D bounding boxes from image space to 3D bounding boxes. Lifting 2D bounding boxes to 3D space can be done with minimal computational overhead leading to real-time capable performance. 

We summarize our contributions as follows:
1) We introduce \netname{} which can be used with all current state-of-the-art 2D object detection frameworks such as SSD \cite{liu2016ssd}, Yolo \cite{redmon2018yolov3} and Faster-RCNN \cite{ren2015faster} to detect vehicles and lift their 2D bounding boxes to 3D space.
2) We exemplary extend SSD with the proposed \netname{} architecture to accentuate the practical use of \netname. 
3) We evaluate \netname{} on 4 challenging and diverse datasets which are especially tailored for autonomous driving: KITTI, nuScenes, A2D2 and Synscapes. We achieve competitive results on the publicly available KITTI 3D Object Detection and nuScenes Object Detection benchmarks. At the same time \netname{} is the fastest 3D object detection framework that relies exclusively on monocular RGB images.

\section{RELATED WORK}
% !TeX root = root.tex

% \cite{MousavianCVPR2017,kim2019inverse,,roddick2018orthographic,siddarth2019learning,xu2018multi,wang2019pseudo,you2019pseudo,ma2019accurate,weng2019monocular,wang2019task}
% \cite{chabot2017deep,kundu20183d,manhardt2019roi,he2019mono3dpp,qin2019monogrnet,barabanau2019monocular,liu2019deep,naiden2019shift,fang20193d,min2019multi,simonelli2019disentangling}
% \cite{ku2019monocular,li2019gs3d,chen2016monocular,brazil2019m3d}
Our work is related to different approaches for monocular 3D object detection. In the following section, we divide the relevant methods into three typical streams of research. First, methods that suggest novel 3D prior bounding box generations or losses. Second, methods that explore geometric reasoning and shape reconstruction. Third, methods that transform input data or feature representations.
\subsection{Novel 3D Prior Bounding Box Proposals or Losses}
Mono3D \cite{chen2016monocular} exhaustively generates possible 3D region proposals based on predictions from class semantics, contour, shape and location priors, leading to slow inference speed. ROI-10D \cite{manhardt2019roi} concatenates features from a 2D Region Proposal Network (RPN) as proposed in \cite{ren2015faster} and depth network in a differentiable Region Of Interest (ROI) lifting layer. These features are used to generate sparse 3D prior boxes, which are optimized w.r.t. their ground truths in 3D space. \cite{ku2019monocular} proposes 3D bounding box priors by utilizing 2D prior bounding boxes for object centroid estimation and predicted point clouds for object scale and shape estimation. M3D-RPN \cite{brazil2019m3d} simultaneously creates 2D and 3D bounding box priors by pre-computing the mean statistics of 3D parameters for each prior individually and learns location-specific features by using 2D convolutions of non-shared weights for joint prediction of 2D and 3D boxes. \cite{simonelli2019disentangling} conducts two-stage 3D object detection by disentangling the dependencies of different parameters. This is achieved by handling groups of parameters individually at loss-level. 

Most relevant to our work is GS3D \cite{li2019gs3d}, where visible surface features are extracted from the 2D image plane and perspectively projected to regular shape. In combination with 2D bounding box features, these features are then used for refining the rough initial 3D bounding box proposals, which are based on the estimation of the class, size and local orientation of 2D bounding boxes. 

Compared with GS3D \cite{li2019gs3d}, our novel keypoint representation serves twofold functions. First, it enhances the initialization of the 3D bounding box by leveraging knowledge from 2D image plane that we use to dynamically determine the prior global orientation and dimension of the 3D bounding box. Moreover, these keypoints are employed as geometric constraints to optimize the 3D bounding box estimation.

\subsection{Geometric Reasoning and Shape Reconstruction}
% \cite{MousavianCVPR2017, barabanau2019monocular, qin2019monogrnet, naiden2019shift, fang20193d, liu2019deep, he2019mono3dpp, gahlert2019beyond, min2019multi} 
Geometry is often employed as a constraint to alleviate the 3D object detection problem. Deep3DBox \cite{MousavianCVPR2017} and Shift R-CNN \cite{naiden2019shift} leverage translation constraints provided by the predicted 2D bounding box by forcing the predicted 3D bounding box to fit tightly within the 2D bounding box using a system of linear equations. In addition to the prediction of dimensions and orientation of the 3D objects, \cite{fang20193d} also classifies the viewpoint and regresses on the center projection of the bottom face of the 3D bounding box, which are used  to jointly fit the projection of 3D bounding boxes to predicted 2D bounding boxes. Mono3D++ \cite{he2019mono3dpp} performs 3D vehicle detection by optimizing both coarse-represented 3D bounding box and fine-grained morphable wireframe. To this end, a ground plane assumption and vehicle shape priors are used.
Instead of per-pixel depth estimation, MonoGRNet \cite{qin2019monogrnet} performs instance-level depth estimation for 3D detection. It explores geometric features to refine the 3D center location of objects by backprojecting the 2D projection of the estimated 3D center to the 3D space based on the estimated depth. \cite{min2019multi} employs the multi-view representation to regress projected 3D bounding boxes into 2D.

In $\text{BS}^3\text{D}$ \cite{gahlert2019beyond} the bounding shape representation, consisting of four coarsely constrained key points, is learned end-to-end for reconstructing 3D bounding box in a post-processing step. 
\cite{barabanau2019monocular} predicts coordinates and a visibility state for 14 predefined 2D keypoints and estimate local orientation based on multi-bin classification. It then learns to estimate correspondences between the 2D keypoints and their 3D counterparts annotated on 3D CAD models for solving the object localization problem. 
DeepManta \cite{chabot2017deep} uses 3D CAD models and annotated 3D parts in a coarse-to-fine localization process, while \cite{kundu20183d} jointly learns the detailed amodal 3D shape and pose by exploring inverse graphics.

In contrast, we select four keypoints directly from the set of 3D bounding box vertices and hence do not rely on keypoints on object level. As a result, we do not depend on any CAD models. These four selected keypoints serve as geometric constraints jointly optimizing the localization, dimensions, and 3D orientation of the 3D bounding box detection.

\begin{figure*}[t!]
\resizebox{\textwidth}{!}{%
  \input{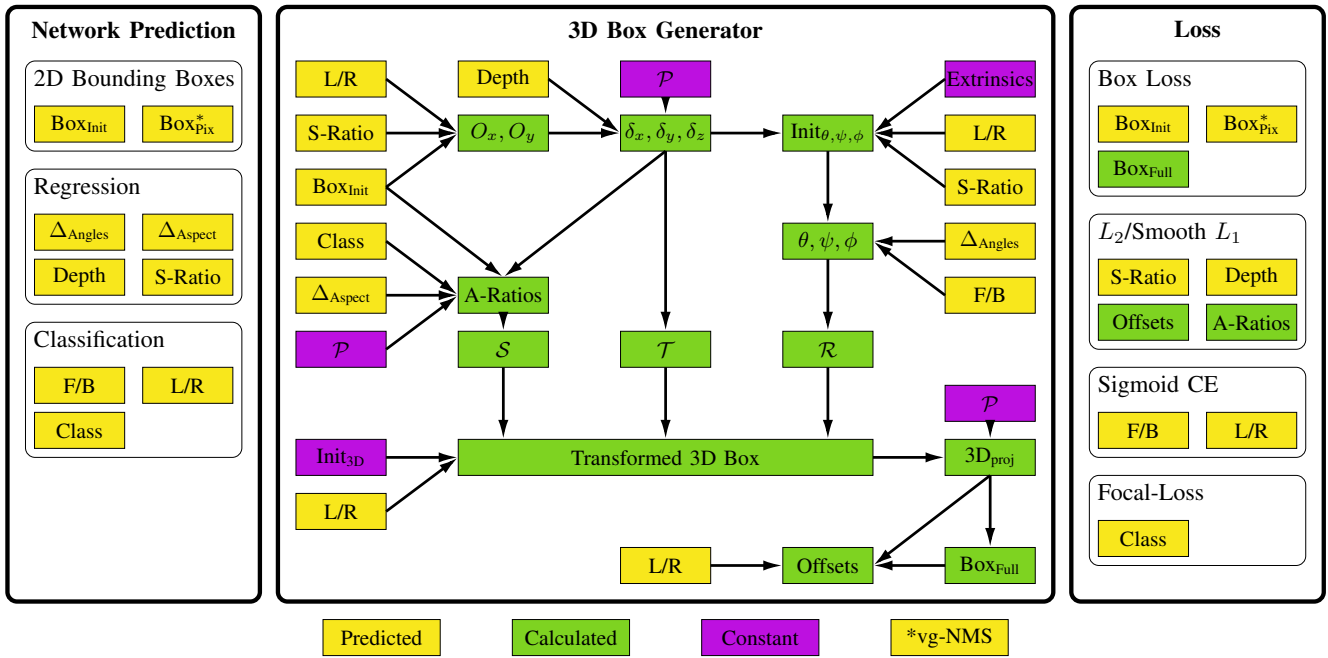}
}%
  \caption{Detailed architecture overview of \netname. Each object represents either a direct network prediction ({\setlength{\fboxsep}{0.5pt}\colorbox{c_yellow}{yellow}}, e.g. Bounding Box predictions and classification parameters), a calculated item ({\setlength{\fboxsep}{1pt}\colorbox{c_green}{green}}, e.g. scaling, translation and rotation matrix $\mathcal S, \mathcal T$ and $ \mathcal R$) or a constant ({\setlength{\fboxsep}{0.5pt}\colorbox{c_purple}{purple}}, e.g. the projection matrix $\mathcal P$ or the camera extrinsics). \textbf{Left:} Different types of explicit network predictions like 2D Bounding Boxes, Regression and Classification parameters. A description of these predictions can be found in \autoref{sub:prediction}. \textbf{Middle:} Graph of the calculation steps performed in the 3D Box Generator. The 3D Box Generator calculates a scaling, translation and rotation matrix $\mathcal S, \mathcal T$ and $ \mathcal R$ for each object. Using this set of transformations the final 3D bounding box is calculated and back-projected into image plane. See \autoref{sub:3dboxcalc} for more details. \textbf{Right:} The losses that are used in \netname. Different types of losses are applied both explicitly on network predictions as well as implicitly on calculated objects as described in \autoref{sub:losses}. Best viewed in color. \label{fig:netoverview}}
\end{figure*}

\subsection{Input Data or Feature Representation Transformation}
3D object detection can be facilitated by directly transforming the input data representation -- such as from monocular RGB images to 3D pseudo point clouds -- or feature representation. \cite{kim2019inverse} restores distance information by projecting the front image onto a bird's-eye-view (BEV) image after correcting the motion change of the ego vehicle by utilizing the extrinsic parameters of the camera and inertial measurement unit information. A one-stage object detector, Yolov3 \cite{redmon2018yolov3}, is then used to determine the class, location, width and height and the orientation of objects. \cite{siddarth2019learning} uses a GAN \cite{goodfellow2014generative} to generate 3D data. \cite{wang2019pseudo, you2019pseudo} focus on generating a pseudo point cloud by using stereo disparity estimation. \cite{ma2019accurate} transforms the estimated depth into a point cloud with the assistance of camera calibration, followed by a multi-modal feature fusion module embedding the complementary RGB cue into the generated point cloud representation. \cite{weng2019monocular} extracts a set of point cloud frustums based on the generated pseudo point clouds and 2D object proposals in the input image, which are passed to train a state-of-the-art two-stage LiDAR 3D bounding box detector \cite{qi2018frustum}. \cite{roddick2018orthographic} introduces an orthographic feature transform operation that populates the 3D voxel feature map with features from the image-based feature map. \cite{xu2018multi} performs multi-level fusion both on feature and data representation for 3D object detection. 

\section{MODEL}
% !TeX root = root.tex

While standard 2D object detection frameworks are solely built upon a classification and a bounding box regression part, several adjustments are required to lift 2D bounding boxes to 3D ones. \autoref{fig:netoverview} illustrates an overview about the additional prediction modules that are used in our network architecture. In addition, it depicts the required steps to generate a 3D bounding box using these predictions and their corresponding loss function.

In general \netname{} follows a 3-step-scheme to lift a 2D detection into 3D:
\begin{enumerate}
\item Estimation of a 3D bounding box initialization.
\item Refinement of the aspect ratios and size.
\item Refinement of the angles of rotation.
\end{enumerate}

\subsection{Coordinate Frame}
We define a coordinate frame for the object of interest with the setup shown in \autoref{fig:coordinate}. $O$ is located on the bottom of the object and is chosen to be the point which is closest to the camera. $A$ is defining the width of the object relative to $O$ while $B$ and $C$ are used to determine the height and the length of the object, respectively.

\subsection{Network Prediction}
\label{sub:prediction}
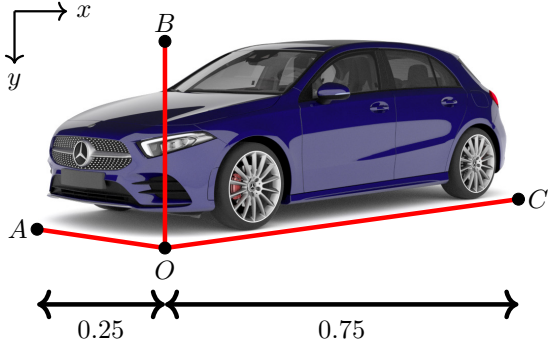
\begin{figure}[t]
\begin{center}
\begin{tikzpicture}
\node[inner sep=0pt] (russell) at (0,0)
    {\includegraphics[width=0.4\textwidth]{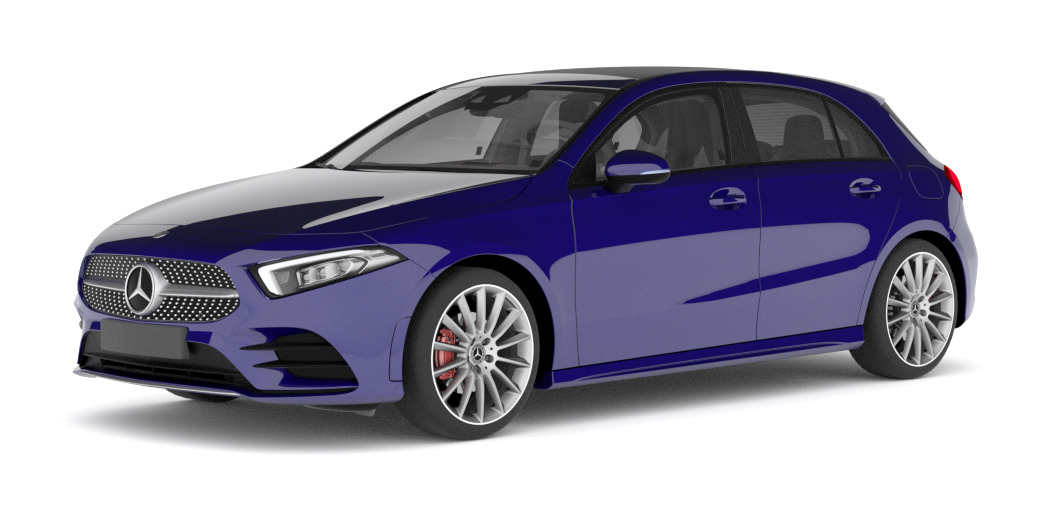}};

\draw[->,black,line width=0.3mm] (-3.5,1.7) -- (-2.8,1.7);
\node[right] at (-2.8,1.7) {$x$};
\draw[->,black,line width=0.3mm] (-3.5,1.7) -- (-3.5,1.0);
\node[below] at (-3.5,1.0) {$y$};

\draw[red,line width=0.6mm] (-1.5,-1.45) -- (-3.2,-1.2);
\draw[red,line width=0.6mm] (-1.5,-1.45) -- (3.2,-0.8);
\draw[red,line width=0.6mm] (-1.5,-1.45) -- (-1.5,1.3);

\draw[<->,black,line width=0.6mm] (-1.5,-2.2) -- (3.2,-2.2);
\draw[<->,black,line width=0.6mm] (-1.5,-2.2) -- (-3.2,-2.2);

\node[below] at (-2.35,-2.3) {$0.25$};
\node[below] at (0.85,-2.3) {$0.75$};

\draw[black,fill=black] (-1.5,-1.45)  circle (.5ex);
\draw[black,fill=black] (-3.2,-1.2)  circle (.5ex);
\draw[black,fill=black] (-1.5,1.3)  circle (.5ex);
\draw[black,fill=black]  (3.2,-0.8)  circle (.5ex);

\node[below] at (-1.5,-1.5) {$O$};
\node[left] at (-3.2,-1.2) {$A$};
\node[above] at (-1.5,1.3){$B$};
\node[right] at (3.2,-0.8){$C$};
\end{tikzpicture}
\end{center}
\caption{The coordinate frame as used in \netname. $O$ is the camera closest point, $A$,$B$ and $C$ define the width, height as well as the length of the 3D bounding box. For this configuration the side-ratio is $0.75$.\label{fig:coordinate}}
\end{figure}

In \netname{} there are 3 different types of prediction modules as shown in \autoref{fig:netoverview} \textbf{(left)}:

\textbf{2D Bounding Boxes:}
$\text{Box}_\text{Init}$ is a standard 4-parameter 2D bounding box regressor that is defined by $A_x, B_y, C_x, O_y$ as shown in \autoref{fig:coordinate}. In addition, $\text{Box}_\text{Pix}$ will be trained to predict only the visible parts of the actual vehicle. However, this bounding box regressor is solely used for selecting the best 2D predictions in the underlying 2D object detection framework using vg-NMS \cite{gaehlert2019vg} and is not mandatory for the actual 3D bounding box generation.

\textbf{Regression:}
In addition to the standard 2D bounding box regression, multiple additional parameters are regressed in \netname. 

The side ratio (S-Ratio) determines the extent to which the side of the vehicle is visible in contrast to the front or the back as shown in \autoref{fig:coordinate} with a value range between 0 and 1. This value is used to generate an initialization for the yaw angle of the actual 3D bounding box. While the initial yaw angle is dynamically calculated, the initial pitch and roll angles are fixed to the camera extrinsics. 
%This assumption is valid as -- except for cities like San Francisco -- most of the streets are flat without any notable slopes. \todo{i would delete the last sentence since this is only the initialization and one of our selling points are that we predict all 3 angles}

In a next step, the initial angles are refined. $\Delta_\text{Angles}$ is a 3 dimensional vector that represents the offsets of these initially calculated 3D orientations given in radians. $\Delta_\text{Aspect}$ is a 2 dimensional vector that is used for adjusting the aspect ratios $w/h$ as well as $l/h$ where $l,h,w$ refer to the object's length, height and width, respectively.

As the 3D bounding box requires a valid depth and \netname{} operates exclusively on monocular RGB images, the depth of the bounding box needs to be estimated as well. To this end, the inverse euclidean distance from the camera to the origin of the vehicle $O$ is estimated.

\begin{figure}[t]
\begin{center}
\includegraphics[width=0.23\textwidth]{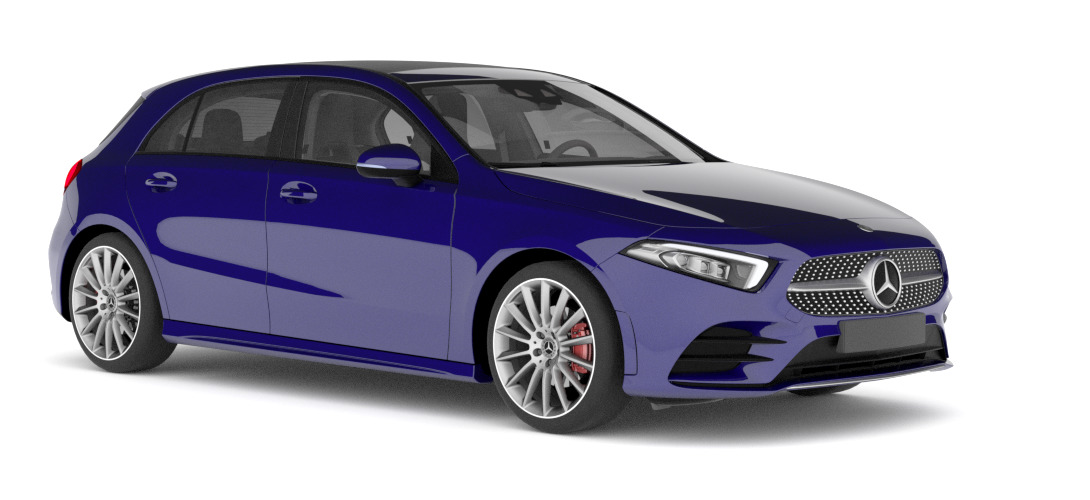}
\includegraphics[width=0.23\textwidth]{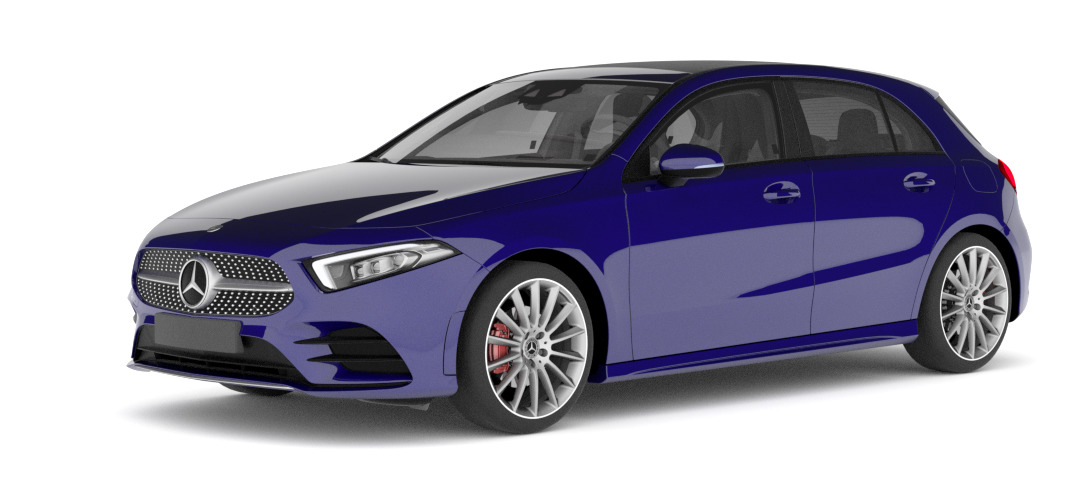}
\includegraphics[width=0.23\textwidth]{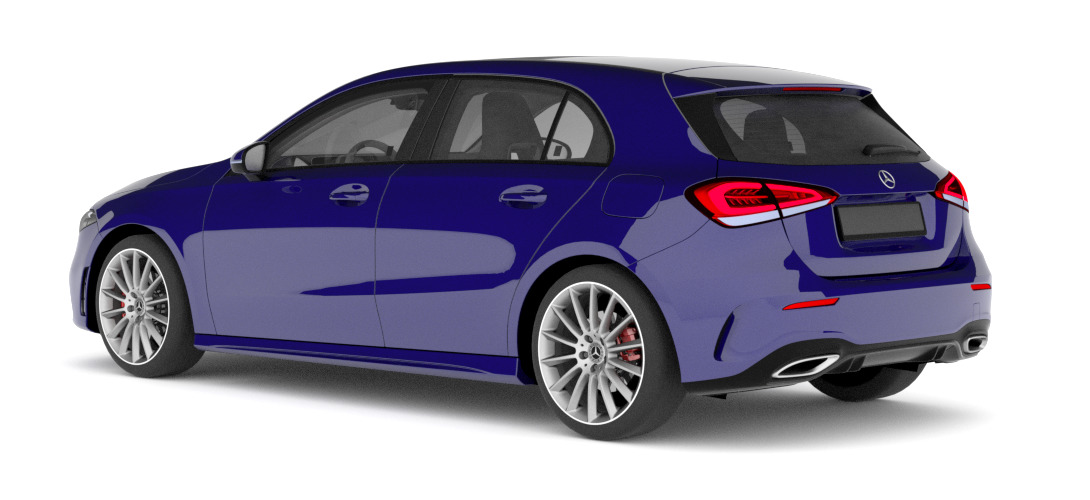}
\includegraphics[width=0.23\textwidth]{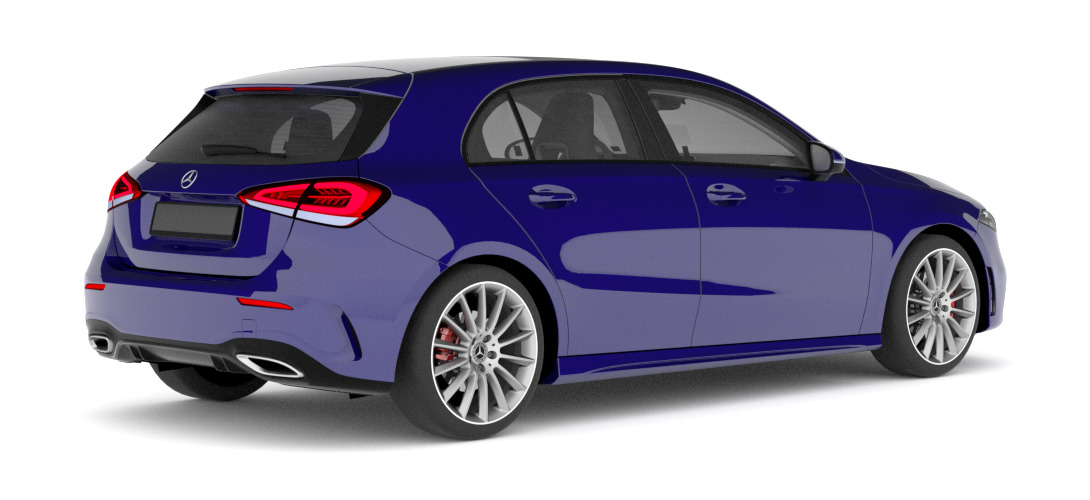}
\end{center}
\caption{Visualization of F/B and L/R classification. The cars in the top have F/B=F as the front is visible. In contrast, for the cars below it is F/B=B. Additionally, both cars on the left are \emph{left} cars with L/R=L. The side of the cars is visible \emph{left} to the front/back while for both cars on the right the sides are visible \emph{right} to front/back and hence L/R=R.\label{fig:lr}}
\end{figure}

\textbf{Classification:}
In addition to the actual object's classification (Class) there are two more classifications performed in \netname. F/B is a binary classification that is trained to predict if the front (F) or the back (B) of the vehicle is visible. This classification simplifies the yaw calculation as the bounding box has to be rotated by $+ \pi$ on the yaw axis depending on the front or the back being visible. Additionally, L/R classifies whether the side of the car is left or right compared to the front or the back. See \autoref{fig:lr} for an example for F/B and L/R classification.

\subsection{3D Box Generator}
\label{sub:3dboxcalc}
In the 3D Box Generator of the \netname{} architecture, the predicted parameters are transformed to a real 3D bounding box for each detection. To this end three matrices are calculated that represent scaling $\mathcal S$, translation $\mathcal T$ and rotation $\mathcal R$. The dependencies for each calculation are illustrated in \autoref{fig:netoverview} \textbf{(middle)}.

\textbf{Translation:} For each $\text{Box}_\text{Init}$ the 2D coordinates of the origin $O$ are calculated as follows:
\begin{align}
O_x &= \begin{cases}
  \text{Box}_{\text{Init},x_\text{min}} + \text{S-Ratio} \times w_\text{2D} & \text{if L/R=L}\\
  \text{Box}_{\text{Init},x_\text{max}}- \text{S-Ratio} \times w_\text{2D} & \text{if L/R=R}
\end{cases} \\
O_y &= \text{Box}_{\text{Init},y_\text{max}}
\end{align} 
with $w_\text{2D}$ being the width of $\text{Box}_\text{Init}$. To determine the 3D position $(\delta_x, \delta_y, \delta_z)$ of $O$, inverse graphics is used under the constraint that the distance of $O_\text{3D}$ is the predicted depth $d$. For inverse graphics the camera projection matrix $\mathcal P$ is needed. Finally, $\mathcal T = \left( \mathds{1} | (\delta_x, \delta_y, \delta_z,1)^T \right)$ is the homogeneous translation matrix.

\textbf{Scaling:} 
As the 3D position of $O$ is known, $\text{Box}_{\text{Init},y_\text{min}}$ can be used to calculate the 3D height $h$:
\begin{align}
\mathcal P \left(\delta_x, \delta_y - h, \delta_z\right)^T\vert_y \stackrel{\mathrm{!}}= \text{Box}_{\text{Init},y_\text{min}} \equiv B_y .\label{eq:scale}
\end{align}
However, \autoref{eq:scale} is an approximation that is only valid for small pitch and roll angles $\psi$ and $\phi$ which is a reasonable assumption for most of the driving scenarios occurring in real world situations.

To calculate both length $l$ and width $w$ of the 3D bounding box, class dependent prior aspect ratios $\text{AR}_\text{Prior}$ are used that are refined using $\Delta_\text{Aspect}$
\begin{align}
l &= \text{AR}_{\text{Prior},l}\left(\text{Class}\right) \times h \times \Delta_{\text{Aspect},l} \\
h &= h \\
w &= \text{AR}_{\text{Prior},w}\left(\text{Class}\right) \times h \times \Delta_{\text{Aspect},w}.
\end{align}
In our experiments we use the prior aspect ratios $\text{AR}_{\text{Prior},l} \equiv l/h = 2.8$ and $\text{AR}_{\text{Prior},w} \equiv w/h = 1.1$ for the class \emph{Car}. 
The homogeneous scaling matrix $\mathcal S$ is finally calculated as $\mathcal S = \mathds{1} \cdot \left(l, h, w, 1\right)^T$.

% for nice layout
\begin{table*}[t!]
\centering
\caption{3D Average Precision (AP) and Average Orientation Score (AOS) for the KITTI \emph{test} set \cite{geiger2012we}. We compare our method with the other approaches that only use monocular RGB images and lift 2D detections to 3D space. We achieve competitive results in both 3D AP and AOS. At the same time \netname{} is the fastest method and capable to run in real-time (FPS $\ge 20$).}
\label{tab:results}
\begin{tabular}{R{2.7cm}|R{1.6cm}R{1.6cm}R{1.6cm}R{1.6cm}R{1.6cm}R{1.6cm} R{1.5cm} }
\toprule
 & \multicolumn{3}{c}{\textbf{3D Average Precision (AP)}} & \multicolumn{3}{c}{\textbf{Average Orientation Similarity (AOS)}} &  \\  
\textbf{Method} & \multicolumn{1}{c}{\textbf{Easy}}  & \multicolumn{1}{c}{\textbf{Moderate}} & \multicolumn{1}{c}{\textbf{Hard}}&\multicolumn{1}{c}{\textbf{Easy}} & \multicolumn{1}{c}{\textbf{Moderate}} & \multicolumn{1}{c}{\textbf{Hard}} & \multicolumn{1}{c}{\textbf{Runtime}}\\ \midrule
%FQNet \cite{liu2019deep}  &{\SI{2.77}{\percent}} & {\SI{1.51}{\percent}}    &{\SI{1.01}{\percent}} & {\SI{93.66}{\percent}} & {\SI{87.49}{\percent}} & {\SI{73.61}{\percent}} & \SI{500}{\milli\second}   \\

GS3D \cite{li2019gs3d}  &\textbf{\SI{4.47}{\percent}} & \textbf{\SI{2.90}{\percent}}    &\textbf{\SI{2.47}{\percent}} & {\SI{85.79}{\percent}} & {\SI{75.63}{\percent}} & {\SI{61.85}{\percent}} & \SI{2000}{\milli\second}   \\

%M3D-RPN \cite{brazil2019m3d}  &\textbf{\SI{14.76}{\percent}} & \textbf{\SI{9.71}{\percent}}    &\textbf{\SI{7.42}{\percent}} & {\SI{88.38}{\percent}} & {\SI{82.81}{\percent}} & {\SI{67.08}{\percent}} & \SI{160}{\milli\second}   \\

%MonoDis \cite{simonelli2019disentangling}  &{\SI{10.37}{\percent}} & {\SI{7.94}{\percent}}    &{\SI{6.40}{\percent}} & -- & -- & -- & \SI{100}{\milli\second}   \\

%MonoFENet \cite{bao2019monofenet}  &{\SI{8.35}{\percent}} & {\SI{5.14}{\percent}}    &{\SI{4.10}{\percent}} & {\SI{91.42}{\percent}} & {\SI{84.09}{\percent}} & {\SI{75.93}{\percent}} & \SI{150}{\milli\second}   \\

%MonoPSR \cite{ku2019monocular}  &{\SI{10.76}{\percent}} & {\SI{7.25}{\percent}}    &{\SI{5.85}{\percent}} & {\SI{93.29}{\percent}} & {\SI{87.45}{\percent}} & {\SI{72.26}{\percent}} & \SI{200}{\milli\second}   \\

%MVRA \cite{min2019multi}  &{\SI{5.19}{\percent}} & {\SI{3.27}{\percent}}    &{\SI{2.49}{\percent}} & \textbf{\SI{95.66}{\percent}} & \textbf{\SI{94.46}{\percent}} & \textbf{\SI{81.74}{\percent}} & \SI{180}{\milli\second}   \\

ROI-10D \cite{manhardt2019roi}  &{\SI{4.32}{\percent}} & {\SI{2.02}{\percent}}    &{\SI{1.46}{\percent}} & {\SI{75.32}{\percent}} & {\SI{68.14}{\percent}} & {\SI{58.98}{\percent}} & \SI{200}{\milli\second}   \\
%Shift R-CNN \cite{naiden2019shift}  &{\SI{6.88}{\percent}} & {\SI{3.87}{\percent}}    &{\SI{2.83}{\percent}} & {\SI{93.75}{\percent}} & {\SI{87.47}{\percent}} & {\SI{77.19}{\percent}} & \SI{250}{\milli\second}   \\

\midrule
\netname{} (ours) & {\SI{3.27}{\percent}} & {\SI{2.52}{\percent}}    &{\SI{2.11}{\percent}} & \textbf{\SI{88.59}{\percent}} & \textbf{\SI{78.44}{\percent}} & \textbf{\SI{66.26}{\percent}} &\textbf{\SI{24}{\milli\second} }  \\
  \bottomrule
\end{tabular}
\end{table*}

\textbf{Rotation:}
Analogous to the scaling values, prior rotation values are estimated that are refined using $\Delta_\text{Angles}$. In an first order approximation the street in front of the ego vehicle is parallel to vehicle coordinate frame. Hence, prior values for pitch $\psi_\text{Init}$ and roll $\phi_\text{Init}$ are set to the inverse of the camera extrinsics -- i.e. if the camera is mounted with a pitch value of \SI{10}{\degree} the prior pitch for the 3D bounding boxes is \SI{-10}{\degree}.

While these assumption are valid for both pitch as well as roll, calculation of the prior value for yaw $\theta_\text{Init}$ directly involves the object of interest. If S-Ratio=0 the object's heading is close to the angle of the camera ray $\theta_\text{Cam}$. If S-Ratio=1, it is close to $\theta_\text{Cam} \pm \tfrac \pi 2$. Hence, it is
\begin{align}
\theta_\text{Init} &= \begin{cases}
  \theta_\text{Cam} +  \arcsin (\text{S-Ratio})  & \text{if L/R=R}\\
  \theta_\text{Cam} -  \arcsin (\text{S-Ratio}) & \text{if L/R=L}
\end{cases} \\ 
\intertext{with} 
 \theta_\text{Cam} &= \arctan \left( \frac {\delta_x}{ \delta_z} \right).
\end{align}
Finally, the prior angles are refined using $\Delta_\text{Angles}$ to determine the actual angles of the 3D bounding box. Furthermore, the yaw angle $\theta$ is corrected by $+ \pi$ in case the front of the object is visible:
\begin{align}
\theta &= \theta_\text{Init} + \begin{cases}
 \Delta_{\text{Angles},\theta} + \pi & \text{if F/B=F}\\
 \Delta_{\text{Angles},\theta} & \text{if F/B=B}
\end{cases}\\
\psi &= \psi_\text{Init} +  \Delta_{\text{Angles},\psi} \\
\phi &= \phi_\text{Init} +  \Delta_{\text{Angles},\phi}.
\end{align}
The joint rotation matrix $R$ is then calculated and the homogeneous rotation matrix $\mathcal R$ computes to $\mathcal R = \begin{pmatrix}
 R & 0\\
 0 & 1
\end{pmatrix}$.

\textbf{3D Bounding Box:} As a last step an initial 3D bounding box with size of \SI{1}{\meter}$\times$\SI{1}{\meter}$\times$\SI{1}{\meter} is transformed using $\mathcal S$, $\mathcal R$ and $\mathcal T$.
This 3D bounding box is projected in the image plane and the $x$- and $y$-offsets for $A$, $B$, $C$ and $O$ of the predicted 3D bounding box are calculated which will be later used in the loss calculation. The offsets are values between $0$ and $1$ and determine the relative position of the keypoints in $\text{Box}_{\text{Full}}$. $\text{Box}_{\text{Full}}$ is the 2D bounding box enclosing the 8 projected vertices of the 3D bounding box.

\subsection{Loss}
\label{sub:losses}
For standard 2D object detection the overall loss is defined as 
\begin{align}
L = \tau  L_\text{cls} + \alpha  L_\text{loc}
\end{align}
with $L_\text{cls}$ and $L_\text{loc}$ denoting the classification and the box localization loss. 

We extend the loss by adding further loss terms corresponding to the predictions and the final bounding boxes as well as the offsets:
\begin{align}
L &= L_\text{boxes} + L_\text{params} + L_\text{class} \\
L_\text{boxes} &= \alpha  L_\text{loc,init} + \beta  L_\text{loc,full} + \gamma  L_\text{loc, pix} \\
L_\text{params} &= \zeta  L_\text{s-ratio} + \eta  L_\text{depth} + \kappa  L_\text{offsets} + \mu  L_\text{a-ratios} \\
L_\text{class} &= \nu  L_\text{F/B} + \xi  L_\text{L/R} + \tau  L_\text{cls}.
\end{align}
For all 2D box regression losses we use $\cos(\text{IoU})$ and for all parameters specific losses we use $L_2$, except for the depth regressor where a smooth $L_1$ loss is used. For $L_\text{F/B}$ and $L_\text{L/R}$, sigmoid cross entropy loss and for the object's classification $L_\text{cls}$ focal loss \cite{lin2017focal} is employed as shown in the network overview \autoref{fig:netoverview} \textbf{(right)}.

\section{EXPERIMENTS}
% !TeX root = root.tex

\subsection{Dataset}
We evaluate \netname{} using several different and challenging datasets tailored especially for autonomous driving purposes. 

KITTI \cite{geiger2013vision} contains approx. \num{15000} images and offers a public benchmark for 3D object detection. However, the images have a rather small resolution compared to more recent datasets like nuScenes \cite{caesar2019nuscenes} or A2D2 \cite{aev2019}. The nuScenes dataset features 1000 distinct scenes in \num{40000} annotated frames and additionally offers a public benchmark for monocular 3D object detection. A2D2 contains \num{12000} annotated frames. 

In contrast to KITTI, nuScenes and A2D2, Synscapes is a synthetic dataset with \num{25000} photo-realistic rendered images. While real world datasets usually only provide annotations for the yaw angle $\theta$, synthetic datasets like Synscapes provide a full description of the 3D object including pitch and roll orientations. As \netname{} is capable of predicting all 3 angles of rotation, we also evaluate on Synscapes. 

Since A2D2 and Synscapes do not offer an official \emph{train}-\emph{val} split and a \emph{test} set, we split the data such that \SI{80}{\percent} of all images are used for training while \SI{20}{\percent} are used for evaluation. We make sure that all images of a recorded sequence are contained either in the \emph{train} or the \emph{val} set.

\subsection{Experimental Setup}
For each dataset we train a single network such that we do not share data of different datasets across the experiments. We use InceptionV1 \cite{googlenet} as the backbone network for our extended \netname{} SSD architecture. While other backbone networks like ResNet \cite{he2016deep} may result in superior detection performances \cite{huang2017speed,bianco2018benchmark}, InceptionV1 offers a good trade-off between speed and accuracy. We use the Adam optimizer with an initial learning rate of $lr=0.0001$. We found that a loss weight of $1$ for all loss weights except for the class and depth weights $\tau = 2$ and $\eta = 0.5$ yields high detection rates and stable training. However, we schedule the training such that the network first focuses on pure 2D SSD box detection and therefore set all weights initially to $0$ except for $\alpha$ and $\tau$. The other weights are logistically increased during the first 100k iterations until they reach their final value. This ensures a high 2D detection performance which is required for a good performance in 3D object detection as \netname{} lifts 2D detections to 3D space. Furthermore, we add a cosine learning rate decay to improve the overall network performance and stability \cite{Loshchilov2017SGDRSG}.

\subsection{Metrics}
3D Average Precision (AP) is used for evaluating the network performance following the official KITTI Object Detection benchmark:
\begin{align}
AP &= \frac 1 {40} \sum_{r \in \lbrace 0, \tfrac 1 {39}, ..., 1\rbrace} p_{interp}(r)\\
p_{interp}(r) &= \underset{\tilde r: \tilde r\ge r}{\max} p(\tilde r). \label{eq:p}
\end{align}
$r = \tfrac {TP}{TP+FN}$ denotes the recall and $p(\tilde r)$ the corresponding precision value for recall $\tilde r$. 
\begin{figure}[t!]
\input{kitti_submission_figure.tex}
\end{figure}
For vehicles, a minimum 3D intersection over union (IoU) of at least \SI{70}{\percent} is required to accept a 3D bounding box as \emph{True Positive} (TP). For a 3D bounding box representing an average car with dimensions $(l/w/h) =  (\SI{4.7}{\meter}/\SI{1.8}{\meter}/\SI{1.4}{\meter})$ a shift of \SI{13}{\centi\meter} in each direction already results in an $\text{IoU} < 0.7$ leading to both a $FP$ and a $FN$. While such an accurate localization estimation is easy to achieve for LiDAR based methods such as \cite{yang2019std,chen2019fast}, it is significantly more difficult for approaches that are solely built upon monocular RGB input as predicting depth from monocular images is a challenging task \cite{roy2016monocular, wu2019spatial}. 

Furthermore, Average Orientation Similarity (AOS) is used to assess the ability of \netname{} to correctly predict the object orientation \cite{geiger2012we}. AOS is calculated similar to AP but uses a normalized cosine similarity $s(r)$ instead of $p(r)$ for $p_{interp}$ in \autoref{eq:p}:
\begin{align}
s(r) &= \frac 1 {\left| \mathcal D(r) \right|} \sum_{i \in \mathcal D(r)} \frac {1+\cos \Delta_\theta^{(i)}} 2 \delta_i
\end{align}
where $\mathcal D(r)$ is the set of all detections at recall $r$.

In addition, we evaluate our approach on the public nuScenes Object Detection benchmark. In contrast to the KITTI benchmark, a prediction is accepted as $TP$ if the 2D center distance $d$ is below a certain threshold $t \in \lbrace 0.5, 1, 2, 4\rbrace$\! m.

\subsection{Results}

 % Runtime on nuscenes: 52 ms

In the KITTI 3D Object Detection benchmark we achieve a 3D AP for class \emph{Car} of \SI{2.52}{\percent} and an AOS of \SI{78.44}{\percent} in the \emph{moderate} category which is competitive to other methods that purely process monocular RGB images and lift 2D detections to 3D space. In general, architectures that enrich and lift 2D detections are comparable fast: \netname{} runs in \SI{24}{\milli\second} ($>$ \SI{40}{\fps}) on an NVidia V100 Tesla GPU without further hardware optimization. \netname{} therefore offers an exceptional trade-off between speed and detection performance as shown in \autoref{fig:fpsvsaos}. While other approaches achieve better detection performance they suffer from a slow inference speed as these methods directly explore in 3D space leading to an increased complexity compared to \netname. 

Quantitative results on the KITTI 3D Object Detection benchmark is listed in \autoref{tab:results}. In the official nuScenes Object Detection benchmark, \netname{} achieves \SI{11.4}{\percent} AP for class \emph{Car}.

We applied the same evaluation metrics as used in KITTI for the nuScenes \emph{val}, A2D2 \emph{val} and Synscapes \emph{val} datasets and obtain similar results as illustrated in \autoref{tab:results_data}. However, we combined the categories \emph{moderate} and \emph{hard} as not all required ground truth information is available for each dataset. \autoref{tab:results_speed} illustrates the runtime of \netname{} for the different datasets. \netname{} is able to run also with more than \SI{20}{\fps} on datasets with high resolution images like nuScenes and A2D2.

\begin{figure*}[t!]
\begin{center}
\includegraphics[trim=50 260 50 0, clip, width=0.7\textwidth]{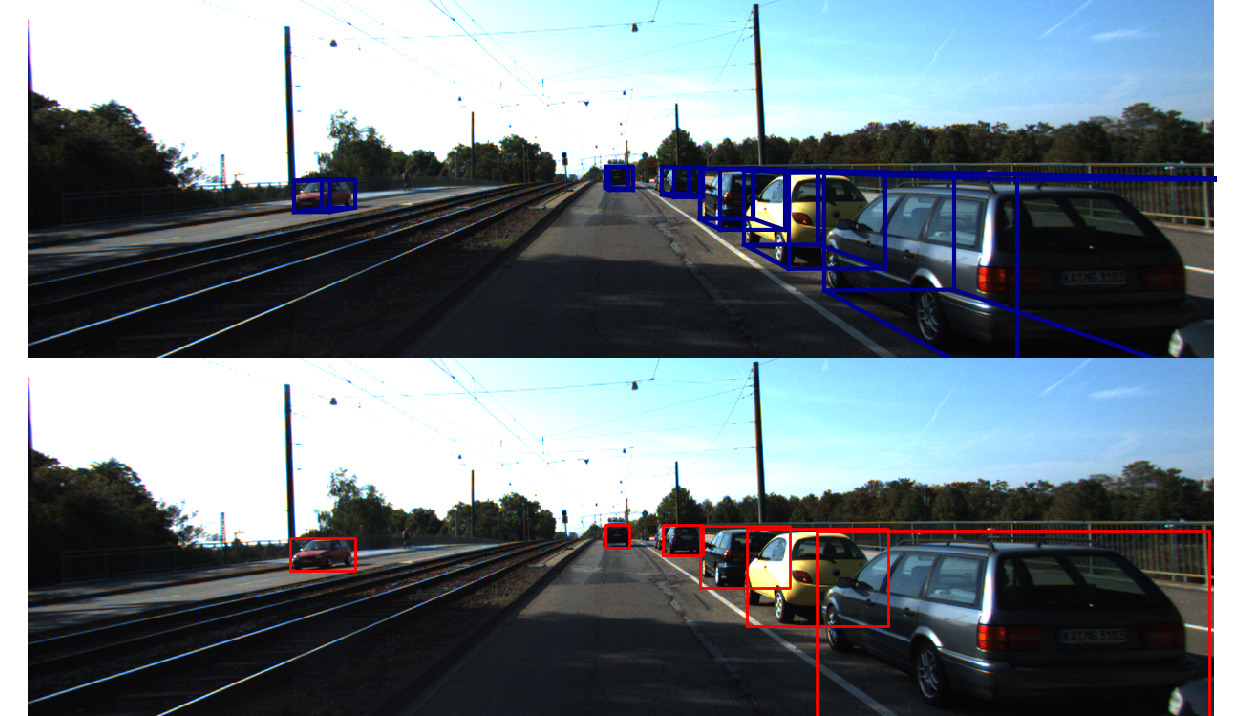} 
\vspace{2mm}
\includegraphics[trim=0 0 0 105, clip, width=0.28\textwidth]{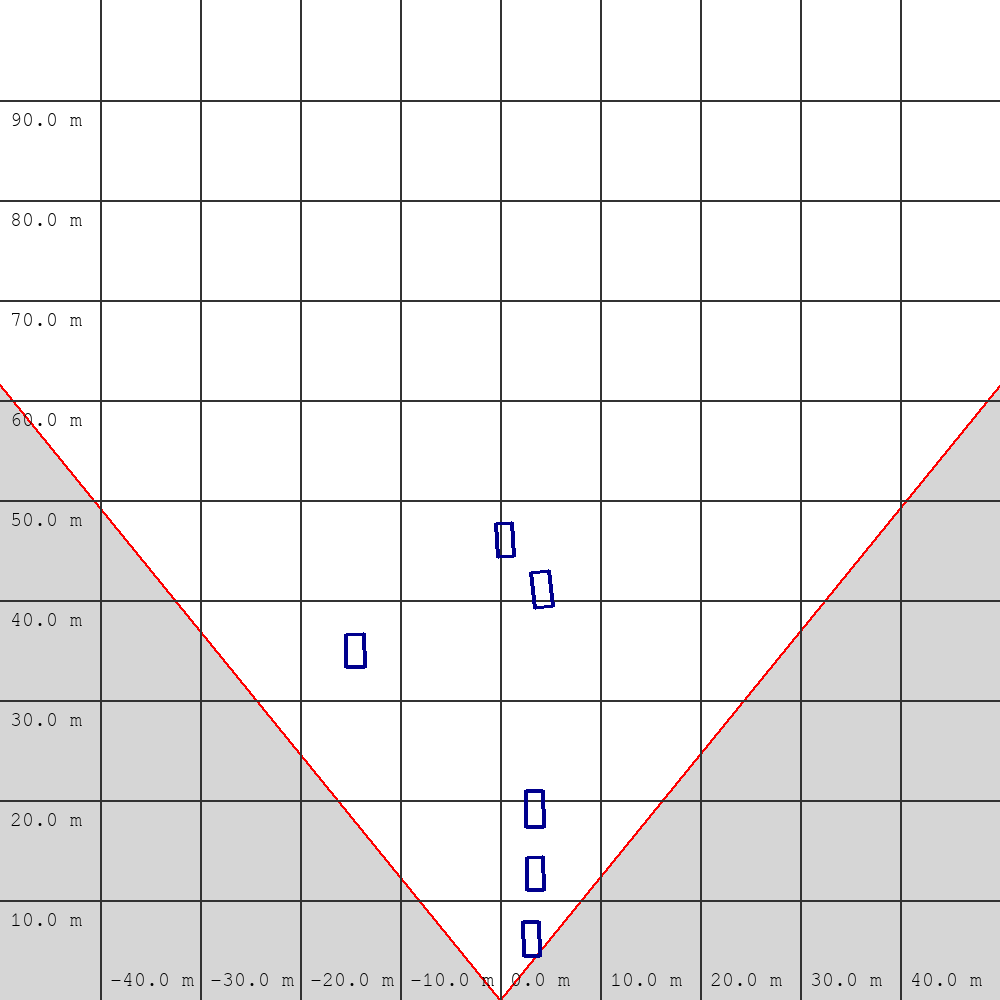}
\vspace{2mm}
\includegraphics[width=0.7\textwidth]{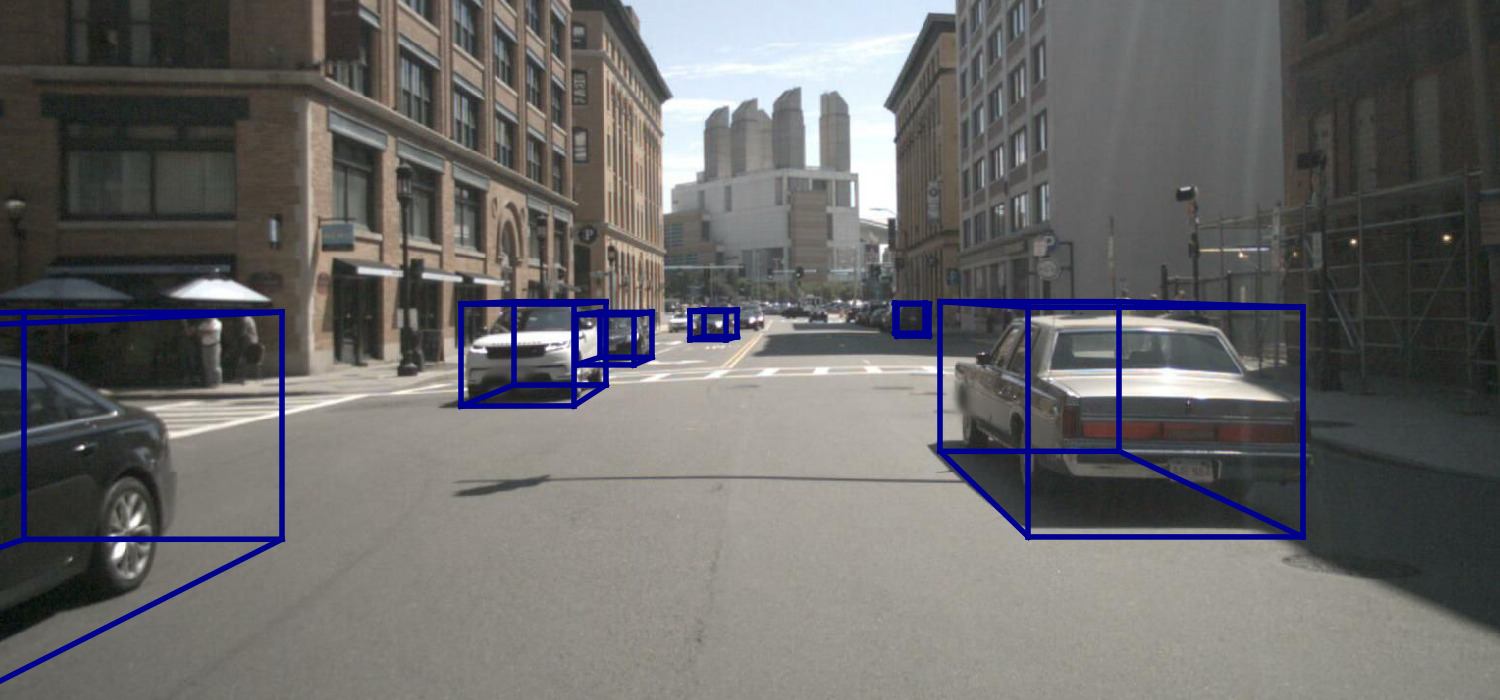} 
\includegraphics[width=0.28\textwidth]{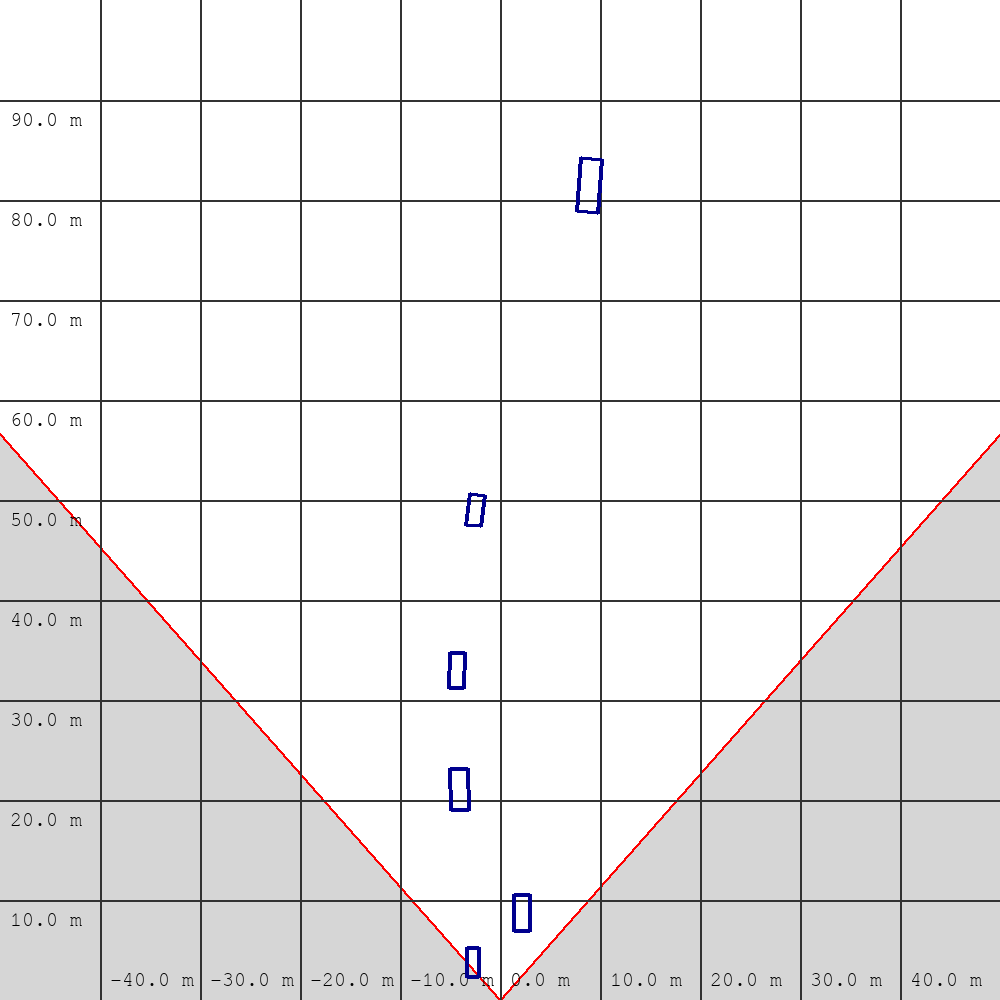}
\vspace{2mm}
\includegraphics[width=0.7\textwidth]{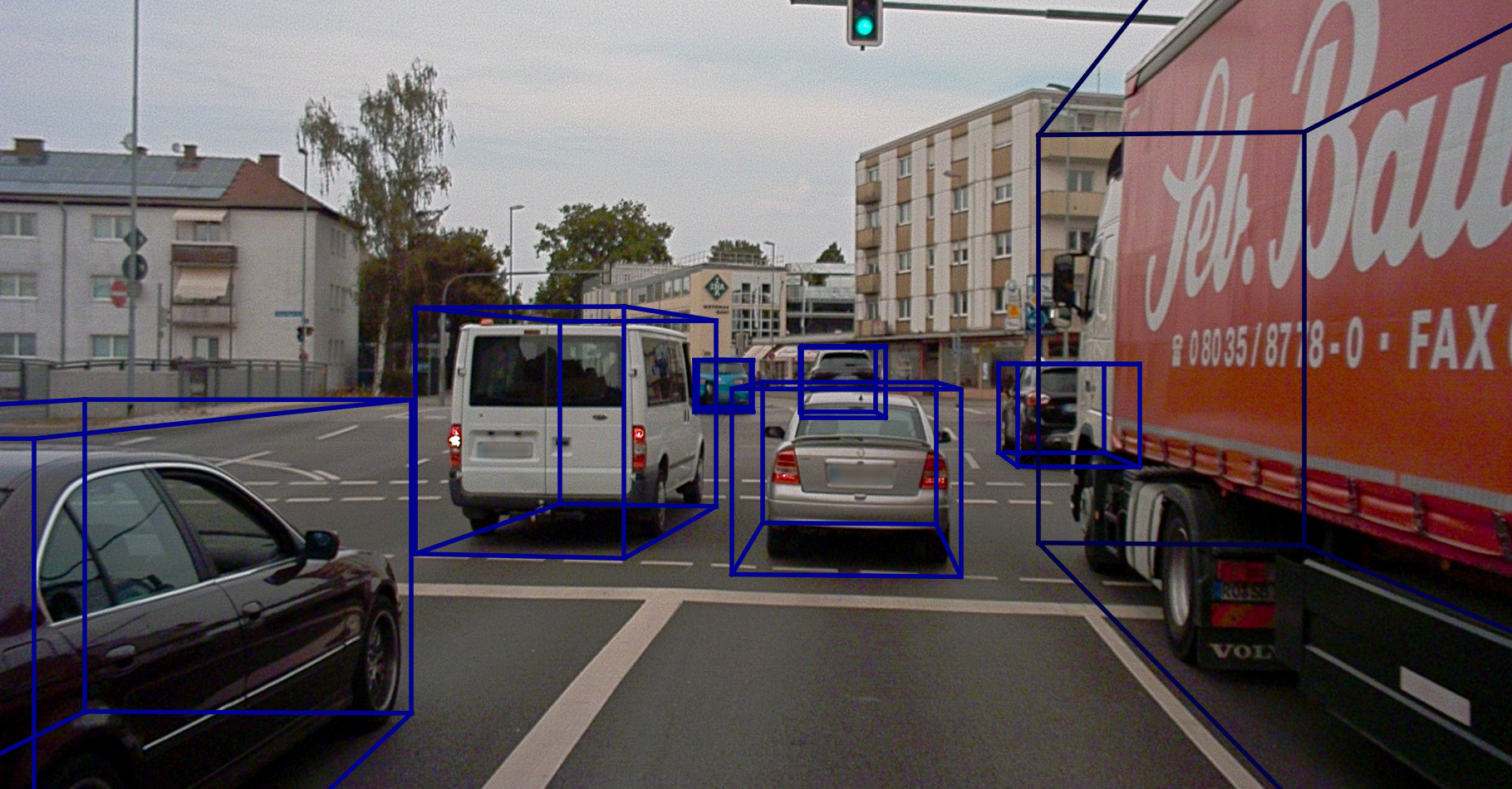} 
\includegraphics[width=0.28\textwidth]{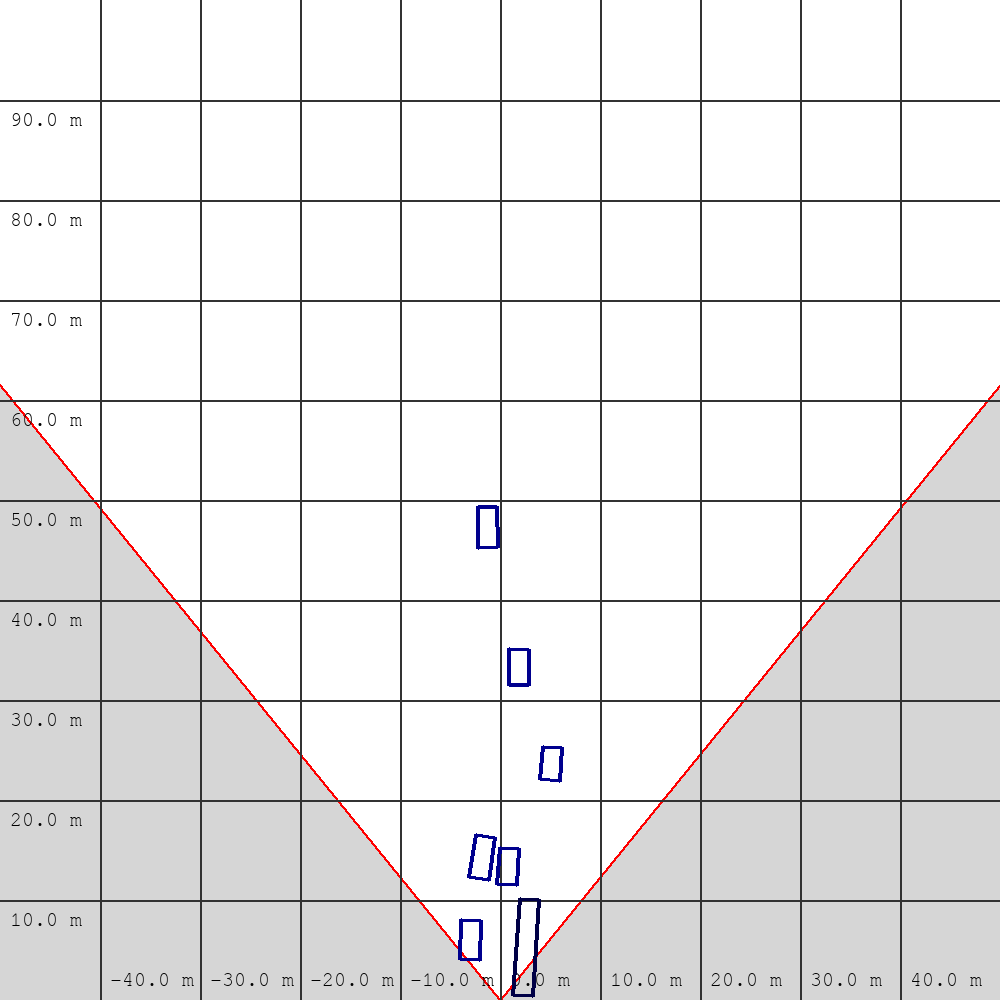}
\includegraphics[width=0.7\textwidth]{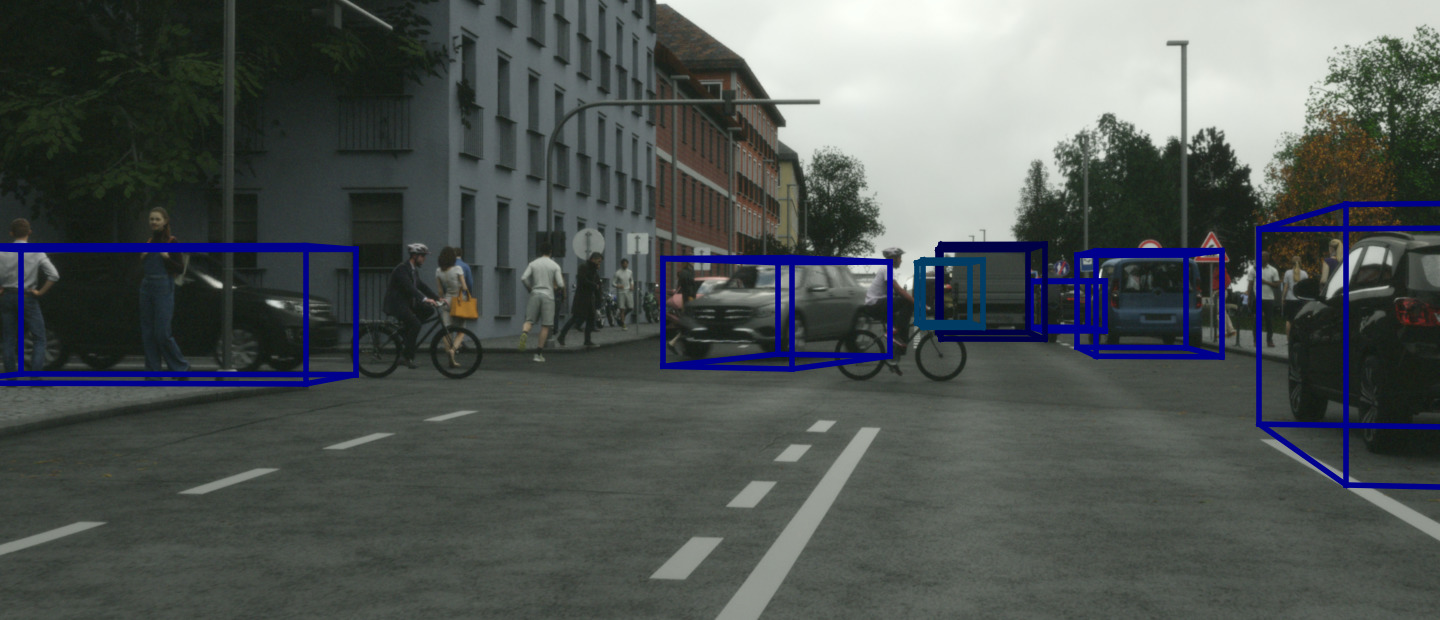} 
\includegraphics[width=0.28\textwidth]{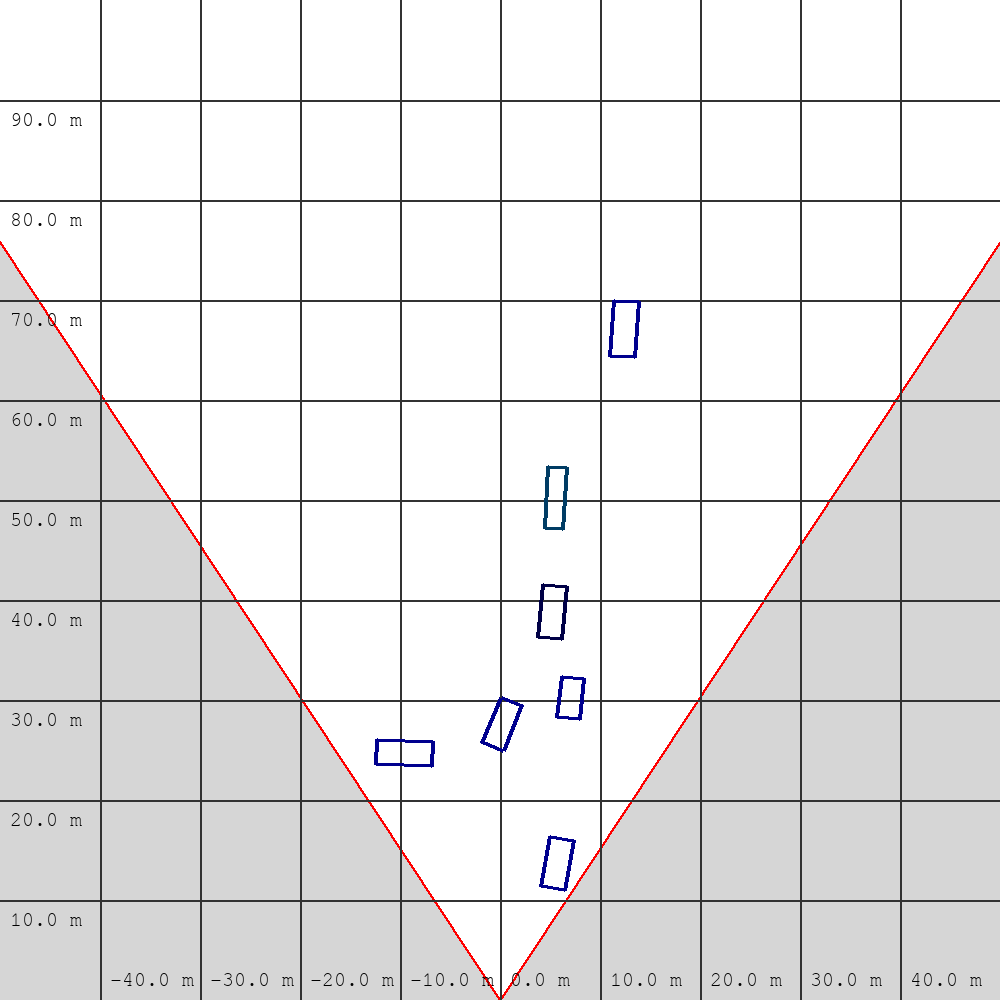}
\end{center}
\caption{Qualitative results for KITTI (\textbf{top}, taken from \emph{test} set), nuScenes (\textbf{2nd row}, from \emph{test} set), A2D2 (\textbf{3rd row}, from \emph{val} set) and Synscapes (\textbf{bottom}, from \emph{val} set). For each image the corresponding bird's-eye-view shown on the right. \label{fig:qualitative}}
\end{figure*}

Qualitative results for each dataset are shown in \autoref{fig:qualitative}. 

\begin{table}[t!]
\centering
\caption{3D Object Detection Results for Different Datasets.}
\label{tab:results_data}
\begin{tabular}{R{2.0cm}|C{0.7cm} R{1.0cm} R{1.cm} R{1.cm} }
\toprule
& &\multicolumn{3}{c}{\textbf{3D Average Precision}} \\ 
\textbf{Dataset} & \textbf{Split} & \multicolumn{1}{c}{\textbf{Easy}} & \multicolumn{1}{c}{\textbf{Moderate}} & \multicolumn{1}{c}{\textbf{Hard}}\\ \midrule
KITTI \cite{geiger2012we} & test &{\SI{3.27}{\percent}} & {\SI{2.52}{\percent}}    &{\SI{2.11}{\percent}}  \\
nuScenes \cite{caesar2019nuscenes} &val & {\SI{1.22}{\percent}} &  \multicolumn{2}{c}{\SI{1.03}{\percent}}     \\ % front only width_1.848
A2D2 \cite{aev2019} & val & {\SI{1.26}{\percent}} &  \multicolumn{2}{c}{\SI{1.26}{\percent}}  \\ %1.88
Synscapes \cite{wrenninge2018synscapes} & val & {\SI{2.51}{\percent}} &  \multicolumn{2}{c}{\SI{2.24}{\percent}}  \\
  \bottomrule
\end{tabular}
\end{table}

\begin{table}[t!]
\centering
\caption{Runtime for several input resolutions. \netname{} is able to be run fast on different resolutions even without hardware specific optimization.}
\label{tab:results_speed}
\begin{tabular}{R{2.cm}|R{3cm} R{1.5cm} }
\toprule
\textbf{Dataset} & \multicolumn{1}{c}{\textbf{Resolution}} & \multicolumn{1}{c}{\textbf{Runtime}}\\ \midrule
KITTI \cite{geiger2012we} &{\SI{1245}{\px}}$\times$\SI{375}{\px} & \SI{24}{\milli\second}   \\
nuScenes \cite{caesar2019nuscenes} &{\SI{1600}{\px}}$\times$\SI{900}{\px} & \SI{43}{\milli\second}   \\
A2D2 \cite{aev2019} &{\SI{1920}{\px}}$\times$\SI{1208}{\px} & \SI{46}{\milli\second}   \\
Synscapes \cite{wrenninge2018synscapes} &{\SI{1440}{\px}}$\times$\SI{720}{\px} & \SI{36}{\milli\second}  \\
  \bottomrule
\end{tabular}
\end{table}

\section{CONCLUSIONS}
% !TeX root = root.tex

In this paper we proposed a novel method for detecting vehicles as 3D bounding boxes for autonomous driving purposes via geometrically constrained keypoints. The basis for this approach is a 2D bounding box detection framework. The 2D detections are then lifted into 3D space. The proposed method can be universally applied to current state-of-the-art 2D object detection frameworks with little computational overhead keeping the runtime close to pure 2D object detection. We added the proposed \netname{} extension to SSD and evaluated our approach on different datasets and benchmarks and achieved competitive results on the challenging KITTI 3D Object Detection and the novel nuScenes Object Detection benchmarks. At the same time our approach is extremely fast and runs with more than \SI{20}{\fps} even on high resolution images without GPU specific hardware optimization. Hence, our approach offers an exceptional trade-off between speed and accuracy and is suitable for being used in productive autonomous systems.

\bibliographystyle{IEEEtran}
\bibliography{bib.bib}
% \bibliography{bib}

\end{document}